\documentclass{ws-us}
\usepackage[sort,compress,super]{cite}
\usepackage{amsmath}
\usepackage{float}
\usepackage{indentfirst}
\usepackage{caption} 
\usepackage[numbers,sort&compress]{natbib}
\usepackage{xurl} 
\usepackage{xcolor}
\usepackage{tabularx}
\usepackage{adjustbox}
\usepackage{algorithm}
\usepackage{algorithmic}
\usepackage{booktabs} 
\usepackage{graphicx}
\usepackage{hyperref}
\begin{document}
\setlength{\pdfpagewidth}{8.5in}
\setlength{\pdfpageheight}{11in}
\catchline{0}{0}{2013}{}{}
\setlength{\parindent}{2em} 
\renewcommand{\topfraction}{0.85}  
\renewcommand{\bottomfraction}{0.65} 
\renewcommand{\textfraction}{0.1}    
\renewcommand{\floatpagefraction}{0.7} 

\setcounter{topnumber}{3}      
\setcounter{bottomnumber}{3}   
\setcounter{totalnumber}{5}    

\markboth{Author's Name}{Paper Title}

\title{UAV-UGV Cooperative Trajectory Optimization and Task Allocation for Medical Rescue Tasks in Post-Disaster Environments}

\author{Kaiyuan Chen$^{a,b}$, Wanpeng Zhao$^c$, Yongxi Liu$^a$, Na Wang$^d$, Yuanqing Xia$^c$, Wannian Liang$^b{^*}$, Shuo Wang$^a{^*}$\footnote{Corresponding authors of this article are Shuo Wang and Wannian Liang. This work is supported by the National Natural Science Foundation of China(No.72441022).}}

\address{$^a$State Key Laboratory of Multimodal Artificial Intelligence Systems, Institute of Automation, Chinese Academy of Sciences,
Beijing, 100190, China
}

\address{$^b$Vanke School of Public Health, Institute for Healthy China, Tsinghua University, Beijing, 100084, China}

\address{$^c$School of Automation, Beijing Institute of Technology, Beijing, 100081, China}

\address{$^d$School of Civil Engineering, Tsinghua University, Beijing, 100084, China}
\address{Email:shuo.wang@ia.ac.cn;liangwn@
 tsinghua.edu.cn}

\maketitle
\begin{abstract}

In post-disaster scenarios, rapid and efficient delivery of medical resources is critical and challenging due to severe damage to infrastructure. To provide an optimized solution, we propose a cooperative trajectory optimization and task allocation framework leveraging unmanned aerial vehicles (UAVs) and unmanned ground vehicles (UGVs). This study integrates a Genetic Algorithm (GA) for efficient task allocation among multiple UAVs and UGVs, and employs an informed-RRT* (Rapidly-exploring Random Tree Star) algorithm for collision-free trajectory generation. Further optimization of task sequencing and path efficiency is conducted using Covariance Matrix Adaptation Evolution Strategy (CMA-ES). Simulation experiments conducted in a realistic post-disaster environment demonstrate that our proposed approach significantly improves the overall efficiency of medical rescue operations compared to traditional strategies. Specifically, our method reduces the total mission completion time to \textbf{26.7 minutes} for a 15-task scenario, outperforming K-Means clustering and random allocation by over \textbf{73\%}. Furthermore, the framework achieves a substantial \textbf{15.1\% reduction in total traveled distance} after CMA-ES optimization. The cooperative utilization of UAVs and UGVs effectively balances their complementary advantages, highlighting the system’s scalability and practicality for real-world deployment.
\end{abstract}


\begin{multicols}{2}
\section{Introduction}

Catastrophic events like earthquakes, floods, and hurricanes frequently destroy critical infrastructure such as roads, bridges, and communication networks, severely hindering traditional emergency response capabilities. In these post-disaster scenarios, providing timely medical assistance becomes an urgent priority, directly impacting survival rates and recovery outcomes. A review of the 2023 Kahramanmaraş earthquakes revealed that significant infrastructure damage impeded rescue efficiency and medical aid, emphasizing the need for better preparedness and more resilient infrastructure\cite{csenol2024emergency}. UAVs and UGVs have emerged as promising solutions due to their flexibility, rapid deployment, and ability to navigate hazardous environments inaccessible to human responders. UAVs offer aerial reconnaissance and swift delivery, bypassing ground-level obstacles, while UGVs provide stable transportation for heavier supplies and possess longer operational endurance.

Significant attention has been given to multi-robot cooperative systems for disaster relief, as they leverage the complementary strengths of UAVs and UGVs. De Castro et al.\cite{machines12030200} investigated UAV-UGV cooperation for coverage tasks in dynamic environments. Krizmancic et al.\cite{8955969} focused on multi-robot coordination for construction tasks, and Cognetti et al.\cite{6942583} utilized UAVs for localization to assist UGV navigation. The relevance of such systems in Synthetic Aperture Radar operations has been reviewed by Queralta et al.\cite{9220149}. 

For task allocation, various techniques, including Genetic Algorithms (GA) , Particle Swarm Optimization (PSO) , and auction-based mechanisms, have been explored. For example, Li et al.\cite{7237695} developed an improved GA for multi-robot task assignment , and Dadgar et al.\cite{DADGAR201662} proposed an adaptive PSO for target searching. Elango et al.\cite{ELANGO20116486} combined K-means clustering with auction mechanisms for balanced task allocation and L{\'o}pez-Gonz{\'a}lez et al.\cite{LOPEZGONZALEZ2020105929} applied parallel GAs for formation control.

Similarly, path planning and optimization methods such as RRT*, Probabilistic Roadmap (PRM), and CMA-ES have been widely used to generate collision-free and efficient trajectories. Li et al.\cite{6640677}, for instance, extended RRT* for multi-robot scenarios to account for motion constraints. Other approaches like the PRM method have been combined with optimization techniques, as seen in the hybrid ABC-PRM-EP framework by Kumar and Sikander\cite{kumar2024novel}. Additionally, trajectory optimization methods such as CMA-ES have been introduced to refine initial paths, minimizing travel distance and overall mission completion time. D{\'i}ez and Aznar\cite{diez2024deep} compared CMA-ES with other evolution strategies for learning autonomous navigation behaviors in robot swarms. Integrating these task allocation and path planning strategies significantly contributes to the success of cooperative multi-robot disaster relief operations.

Despite these notable advancements, several critical limitations in existing research remain unresolved, presenting key research gaps that this study aims to address.

First, most current cooperative frameworks primarily consider homogeneous teams of unmanned vehicles (either UAVs or UGVs alone). This approach ignores the significant benefits of heterogeneous cooperation and fails to fully integrate and exploit the distinct, complementary strengths of UAVs and UGVs, such as a UAV's agility and a UGV's payload capacity.

Second, many existing task allocation methodologies assume static environments and predetermined task sets. This simplification does not reflect the unpredictable nature of real-world disaster scenarios, where new tasks may emerge and the environment is constantly changing. Traditional optimization and heuristic methods often struggle with the high computational complexity of these dynamic, real-time, multi-task scenarios.

Third, trajectory optimization in disaster environments often focuses exclusively on generating feasible paths or minimizing path length, without adequately considering practical constraints like smoothness, continuity, or energy efficiency. Consequently, existing path planning algorithms can produce jagged or inefficient trajectories that are impractical for real-world navigation.

To bridge these gaps, this paper proposes a novel integrated framework explicitly designed for cooperative UAV-UGV operations targeting medical rescue missions in complex post-disaster environments. Our key contributions are summarized as follows:
\begin{itemize}
\item We propose an enhanced Genetic Algorithm  for dynamic task allocation among a heterogeneous fleet, which is a significant improvement over methods focused on homogeneous teams. The allocation is seamlessly integrated with a trajectory generation process that uses Informed-RRT* for initial pathfinding and CMA-ES for sequence and path optimization. This unified approach ensures that vehicle capabilities are matched to tasks and that the resulting paths are globally coordinated and locally optimized.

\item We introduce a robust methodology that not only generates collision-free paths but also refines them for real-world application. The combination of Informed-RRT* and CMA-ES specifically addresses the limitations of traditional algorithms by optimizing task visiting sequences to reduce total travel distance and mission time. A systematic trajectory post-processing approach is also applied to ensure practical smoothness and continuity, addressing a common shortcoming in existing planning frameworks.

\item Through extensive simulation experiments in realistic post-disaster scenarios, we validate the effectiveness, robustness, and scalability of our proposed framework. The selection of a hybrid evolutionary algorithmic framework (EGA and CMA-ES) over alternatives such as exact solvers (e.g., Gurobi) or deep reinforcement learning (e.g., A2C) is motivated by the need for a balance between solution quality and computational efficiency in dynamic, large-scale disaster scenarios. Exact solvers often suffer from prohibitive computational complexity for real-time applications, while reinforcement learning methods require extensive training and exhibit lower interpretability.  We demonstrate significant reductions in total mission completion time and traveled distance compared to traditional strategies like random assignment and K-Means clustering. Critically, our scalability analysis confirms that the proposed approach maintains its efficiency even as the number of tasks and vehicles increases, outperforming baseline methods that show performance degradation under increased complexity.
\end{itemize}

The remainder of this paper is organized as follows: Section 2 provides a detailed literature review. Section 3 formulates the problem. Section 4 presents our integrated methodology. Section 5 details the simulation results and analysis. Finally, Section 6 concludes the paper and discusses future work.

\section{Literature Review}
\subsection{Multi-Robot Cooperation in Disaster Relief}
Multi-robot systems, particularly heterogeneous teams, have emerged as a promising solution for disaster relief operations due to their flexibility, rapid deployment capabilities, and ability to navigate hazardous or inaccessible environments. UAVs excel in aerial reconnaissance and swift delivery, bypassing ground-level obstacles. In contrast, UGVs provide stable transportation for heavier supplies and possess longer operational endurance. The effective cooperation between these platforms leverages their complementary strengths to enhance rescue efficiency and improve survival rates.

Existing studies have explored multi-robot cooperation in various contexts. For example, de Castro et al. investigated UAV-UGV cooperation for coverage tasks in dynamic environments. Krizmancic et al. focused on coordination for automated construction tasks , while Cognetti et al. utilized UAVs for localization to aid UGV navigation. However, a critical limitation in many current frameworks is the focus on homogeneous teams, which fails to fully integrate and exploit the distinct advantages of both UAVs and UGVs. Addressing this gap by developing a framework that can effectively coordinate a heterogeneous fleet is essential for enhancing rescue capabilities.
\subsection{Task Allocation Strategies for Multi-Robot Systems}
Effective task allocation is a crucial challenge in disaster scenarios, as it directly impacts resource deployment and overall mission efficiency. Various optimization and heuristic techniques have been extensively studied for this purpose, including Genetic Algorithms, Particle Swarm Optimization, and auction-based mechanisms. For instance, Li et al. developed an improved GA for multi-robot task assignment , while López-González et al. applied parallel GAs for formation control. Dadgar et al. proposed an adaptive PSO for target searching , and Elango et al. combined K-means clustering with auction mechanisms for balanced task allocation.

The success of these methods highlights the potential of optimization techniques in solving complex, dynamic problems. Beyond traditional heuristics, advanced machine learning and AI methods have also demonstrated significant potential in modeling and forecasting complex patterns in other domains, which underscores their broad applicability. For instance, neural networks have been successfully used for forecasting thermal coal futures trading volumes, peanut oil price changes, corn cash prices, and house prices\citep{xu2022commodity,article11,XU2022200061}. These models often lead to high accuracy and stability, contributing to efficient and low-cost estimations\citep{article1, article2,XU2021106120,XU2021200052,article9}. Similarly, Gaussian process regressions have been effectively applied to predict residential property price indices and steel price indices, often outperforming traditional econometric models\citep{article3,article4,article6,article7,article8}. The use of machine learning has also been extended to material science, such as machine learning the magnetocaloric effect in manganites from lattice parameters\citep{article5}. While these applications are distinct from task allocation, they demonstrate the power of AI/ML models to solve complex, non-linear problems, a critical characteristic of multi-robot task assignment.

However, a key limitation of many existing task allocation methodologies is the assumption of static environments and predetermined tasks, which is unrealistic for dynamic post-disaster situations. This gap necessitates the development of more robust optimization approaches that can handle the high computational complexity of real-time, dynamic, and multi-task scenarios.
\subsection{Trajectory Planning and Optimization Techniques}

Path planning is another crucial aspect of multi-robot missions, requiring the generation of collision-free and efficient trajectories in complex, cluttered environments. Algorithms such as RRT*, Probabilistic Roadmap (PRM), and A* search have been widely employed for this purpose. For example, Li et al. extended RRT* for multi-robot scenarios, considering motion constraints. Other hybrid approaches, like the ABC-PRM-EP framework, have combined PRM with optimization techniques to improve efficiency.

Furthermore, trajectory optimization methods are used to refine initial paths and minimize travel distance and mission completion time. For example, CMA-ES has been compared with other evolution strategies for learning autonomous navigation behaviors in robot swarms. The need for advanced optimization is also evident in other fields where complex data patterns require sophisticated modeling. Graph models, for instance, have been used to analyze causal orderings among property and steel product prices, revealing complex dynamics. Similarly, integrated or composite methods, which combine multiple models, have been shown to improve forecasting accuracy for corn cash prices\citep{xu2022commodity,article11,XU2022200061}.

Despite notable advancements in multi-robot cooperation, task allocation, and trajectory optimization, several critical limitations remain unresolved. 

First, most existing cooperative frameworks primarily focus on homogeneous teams of unmanned vehicles, neglecting the substantial benefits derived from a heterogeneous UAV-UGV partnership. The distinct, complementary strengths of UAVs (agility and rapid coverage) and UGVs (enhanced payload capacity and longer endurance) are not yet sufficiently integrated or exploited in current frameworks.

Second, many proposed task allocation methodologies rely on assumptions of static environments and predetermined task sets, which fail to capture the dynamic and unpredictable nature of real-world post-disaster scenarios. Traditional optimization and heuristic methods often struggle with the high computational complexity that arises in these dynamic, multi-task environments. 

Third, trajectory optimization for disaster environments frequently prioritizes feasibility or path length minimization without adequately considering practical constraints such as smoothness, continuity, or energy efficiency, which are crucial for actual UAV and UGV operations. As a result, existing planning algorithms often produce jagged or inefficient trajectories that are impractical for real-world deployment.

\section{Problem Formulation}
Before delving into the specific mathematical formulations, it is crucial to articulate the overarching problem-solving paradigm that guides our approach. The integrated problem of task allocation, trajectory generation, and sequence optimization for a heterogeneous team in a complex environment is inherently high-dimensional and computationally intractable if solved monolithically. To address this challenge, we adopt a structured, hierarchical, and iterative optimization framework, decomposing the complex problem into a series of interdependent sub-problems.

The process begins with a global task allocation using an EGA, which efficiently distributes tasks among the heterogeneous fleet of UAVs and UGVs by considering their distinct capabilities and constraints. This step produces a preliminary assignment that balances the workload. Subsequently, for each vehicle, the Informed-RRT* algorithm is employed to generate an initial, collision-free, and dynamically feasible trajectory that connects its assigned tasks, transforming the abstract assignment into concrete, safe paths. Finally, to refine the efficiency of these paths, the CMA-ES is applied to optimize the visiting sequence of the assigned tasks for each vehicle, minimizing travel distance and time.

This "Allocate-Plan-Optimize" workflow provides a tractable and effective strategy. The EGA ensures a globally intelligent distribution of work, Informed-RRT* guarantees local feasibility and safety, and CMA-ES performs a fine-grained optimization to eliminate inefficiencies. The outputs of each stage inform and constrain the next, creating a coherent solution pipeline.

This section establishes the formal basis for the cooperative UAV-UGV medical rescue problem addressed in this study. We depict the representation of the post-disaster operational environment, detail the specific capabilities and limitations inherent to the unmanned aerial and ground vehicles employed, define the characteristics of the required rescue tasks, and outline the primary optimization objectives that guide the planning and execution of the mission.
\subsection{Environment Modeling}

The operational theatre is conceptualized as a bounded two-dimensional Euclidean space, denoted by $\mathcal{E} \subset \mathbb{R}^2$, which represents a disaster-impacted zone, typically an urban or semi-urban landscape significantly damaged by events such as earthquakes or widespread flooding. Within this environment $\mathcal{E}$, several key elements are defined. A primary consideration is the presence of $N$ static obstacles, collected in the set $\mathcal{O} = \{O_1, O_2, \dots, O_N\}$. These obstacles, geometrically represented (e.g., as polygons or circles), correspond to impassable areas like collapsed structures, extensive debris fields, or flooded regions that impede vehicle movement. The locations and geometries of these major obstacles are assumed to be known or rapidly mappable post-disaster. Distributed across the environment are $M$ discrete task locations, $\mathcal{T} = \{T_1, T_2, \dots, T_M\}$, each signifying a point where a specific medical rescue action is required, such as the delivery of critical supplies or site assessment. For the scope of this work, these task locations are assumed to remain fixed throughout the planning phase. Finally, a central base station $B$ is situated within $\mathcal{E}$, serving as the designated origin and final destination for all unmanned vehicles involved in the operation, functioning as the mission control and logistics hub. Consequently, the navigable area for the vehicles, or free space, is $\mathcal{E}_{free} = \mathcal{E} \setminus \bigcup_{i=1}^{N} O_i$. A fundamental requirement is that all vehicle movement must be confined to $\mathcal{E}_{free}$, ensuring mandatory collision avoidance with respect to all obstacles in $O_i \in \mathcal{O}$. While real-world disaster zones are inherently dynamic, this study adopts a static representation of major obstacles and task points for planning tractability during a given mission cycle. The schematic diagram is shown in Fig. 1.
\begin{figure}[H]
    \centering
    \includegraphics[width=1\linewidth]{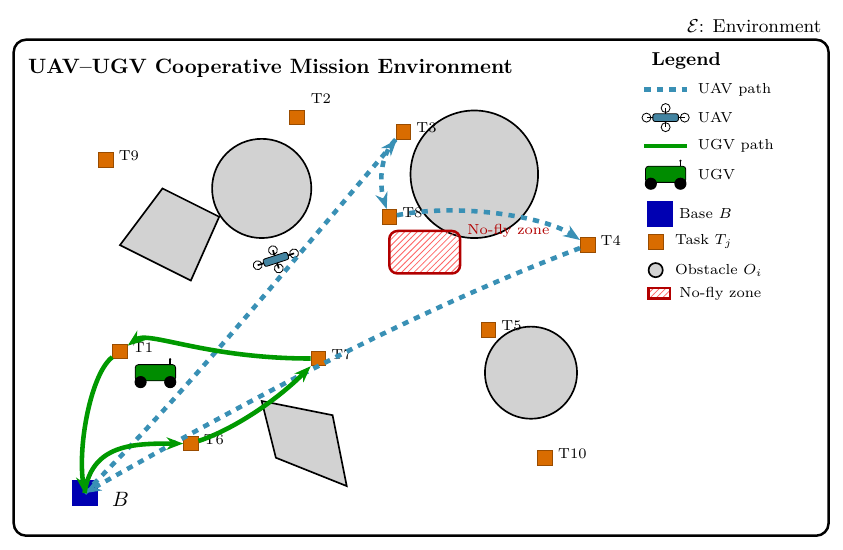}
    \caption{Schematic diagram of the UAV-UGV collaborative mission environment}
    \label{aaa}
\end{figure}

\subsection{UAV and UGV Operational Constraints}
The system utilizes a heterogeneous fleet $\mathcal{V}$ UAVs, forming the set $\mathcal{U} = \{U_1, \dots, U_{N_u}\}$, and UGVs, forming the set $\mathcal{G} = \{G_1, \dots, G_{N_g}\}$. These vehicle types possess distinct operational profiles and are subject to different constraints, which must be explicitly considered for realistic planning.

UAVs offer significant mobility advantages due to their three-dimensional flight capability, although planning often simplifies this to 2.5D or projects trajectories onto the 2D plane $\mathcal{E}$ when altitude primarily serves obstacle negotiation. Their high speed and agility allow them to potentially bypass smaller ground obstacles, though they must avoid larger designated hazards and no-fly zones. However, UAVs are constrained by a limited payload capacity $W_{UAV,max}$ and finite endurance, typically dictated by battery life or fuel, imposing a maximum flight distance $D_{UAV,max}$ or operational time $T_{UAV,max}$. Thus, any planned UAV trajectory must satisfy:
\begin{equation}
\sum_{i=0}^{n-1} d(P_i, P_{i+1}) \le D_{UAV,max},
\end{equation}
where $P_i$ denotes the \textbf{waypoints} along the planned trajectory generated by the path planning algorithm (e.g., Informed-RRT*), $d(\cdot, \cdot)$ is the Euclidean distance, and $D_{\text{UAV,max}}$ is the maximum flight distance.

UGVs, conversely, operate exclusively on the ground surface. Their maneuverability is restricted as they must navigate entirely around obstacles defined in $\mathcal{O}$, generally at lower speeds than UAVs. Their primary advantages lie in a typically higher payload capacity $W_{UGV,max}$, making them suitable for transporting heavier supplies, and greater operational endurance, characterized by a maximum travel distance $D_{UGV,max}$ or time $T_{UGV,max}$. Similar to UAVs, their total path length is constrained: 
\begin{equation}
\sum_{i=0}^{n-1} d(P_i, P_{i+1}) \le D_{UGV,max},
\end{equation}
 where Pi are waypoints, $d( , )$ denotes the Euclidean distance.UGVs are also subject to kinematic constraints, potentially including non-holonomic restrictions, and terrain traversability limitations, alongside the fundamental requirement to avoid collisions with obstacles and other vehicles.

While this constraint is formulated using the sum of Euclidean distances for simplicity and computational tractability, it is acknowledged that the actual energy consumption of UAVs and UGVs is more accurately represented by the total length of the final, dynamically feasible trajectory, which incorporates the effects of path curvature, kinematic constraints, and environmental factors. The total trajectory length serves as a better proxy for real-world energy expenditure than the sum of straight-line segments. In future work, we aim to integrate more sophisticated energy models that account for vehicle dynamics and terrain characteristics to further refine the planning process.

The successful deployment of this heterogeneous system hinges on effectively leveraging the complementary strengths of each vehicle type – the speed and access of UAVs balanced against the payload and endurance of UGVs. Coordination strategies aim to assign tasks intelligently and ensure efficient, conflict-free execution within the shared operational space, respecting all individual vehicle constraints. This coordination implicitly assumes reliable communication capabilities between vehicles and the base station for command, control, and information sharing.

\subsection{Task Definition and Constraints}
The central aim of the mission is the successful servicing of the set of defined medical rescue tasks $\mathcal{T} = \{T_1, T_2, \dots, T_M\}$, each located at a specific point $p_j \in \mathcal{E}$. The allocation and execution of these tasks are governed by several critical constraints. Primarily, each task $T_{j}$ must be assigned to and completed by exactly one vehicle $V_i \in \mathcal{V}$; this exclusive service constraint can be represented using a binary assignment variable $x_{ij}$ (binary variable) such that $\sum_{i=1}^{|\mathcal{V}|} x_{ij} = 1$ for every task j. While this study assumes broad accessibility, capability matching may further restrict assignments in scenarios where specific tasks are only feasible for aerial or ground platforms. Furthermore, vehicles must respect their resource limitations; the cumulative requirements (e.g., payload weight/volume) of tasks assigned to a vehicle cannot exceed its capacity ($W_{UAV,max}$ or $W_{UGV,max}$). Crucially, the total path length necessary for a vehicle $V_{i}$ to depart from the base $B$, visit its assigned tasks $\mathcal{T}_i \subseteq \mathcal{T}$ in sequence, and return to B, must not surpass its operational endurance limit . The requirement for each vehicle to return to the base station completes the mission cycle. For simplification, this work assumes tasks are independent, without precedence constraints or differing priorities, and lack strict completion time windows, although minimizing overall mission duration remains a core objective.

\subsection{Optimization Objectives}
The overarching goal of the proposed framework is to determine an optimal allocation of tasks among the available UAVs and UGVs, and to subsequently compute efficient, safe, and feasible trajectories for each vehicle to execute its assigned duties. The effectiveness of the rescue operation is primarily evaluated against several optimization objectives. A key objective is the minimization of the total mission completion time, or makespan ($T_{makespan}$), defined as the time elapsed until the last vehicle finishes its assigned tour and returns to the base station. Mathematically, this is expressed as 
\begin{equation}
    \min T_{makespan} = \min \left( \max_{V_i \in \mathcal{V}} \{ T_{mission,i} \} \right),
\end{equation}
where $T_{mission,i}$ represents the total duration of vehicle $V_{i}$ mission. Another significant objective is to minimize the total travel distance ($L_{total}$) accumulated across all vehicles in the fleet, formulated as 
\begin{equation}
    \min L_{total} = \min \left( \sum_{V_i \in \mathcal{V}} L_i \right),
\end{equation}
where $L_{i}$ is the path length for vehicle $V_{i}$. This objective directly correlates with overall energy consumption and operational wear. While optimizing time and distance is crucial, it is imperative that all generated plans strictly adhere to safety and feasibility requirements; planned trajectories must be collision-free with respect to both static obstacles $\mathcal{O}$ and other moving vehicles, and must satisfy all vehicle-specific operational constraints (endurance, payload, kinematics). Recognizing that minimizing makespan and total distance can sometimes be conflicting goals, this work often addresses the multi-objective nature through methods like weighted sum optimization or by prioritizing one objective while treating others as constraints, always enforcing safety and feasibility as hard constraints. The problem, therefore, constitutes a complex multi-vehicle task allocation and trajectory optimization challenge set within a constrained and hazardous environment.
\subsection{Problem Scope and Formal Summary}
In essence, the problem tackled in this paper is the optimization of task allocation and trajectory planning for a heterogeneous fleet $\mathcal{V}$ of UAVs and UGVs operating in a post-disaster setting. The operational context is defined by an environment $\mathcal{E}$ containing a set of known, static obstacles $\mathcal{O}$, a designated set of medical rescue task locations $\mathcal{T}$, and a central base station $B$. The core objective is to determine an optimal operational plan, which encompasses both the assignment of each task $T_j \in \mathcal{T}$ exclusively to a single vehicle $V_i \in \mathcal{V}$ (represented by assignment $X$), and the subsequent generation of an ordered task sequence $\pi_i$ and a feasible, collision-free trajectory $\mathcal{P}_i(t)$ for each vehicle. Each trajectory must originate from $B$, visit the assigned tasks in the determined sequence $\pi_i$, and terminate back at $B$, while strictly adhering to all vehicle-specific constraints, particularly maximum travel endurance and payload limits. The quality of the plan is primarily evaluated based on minimizing the overall mission completion time and the total distance traveled by the entire fleet. Critically, the return leg from the final task location back to the base station $B$ is planned as a distinct, independent path segment using the Informed-RRT* algorithm, ensuring an optimal and collision-free route, rather than simply reversing the outbound path.

It is important to recognize that this problem formulation, for tractability and focus, operates under several key simplifying assumptions. These include the static nature of the environment (obstacles and tasks do not change during the mission), the availability of perfect information regarding the environment and vehicle states, idealized vehicle control, and the assumption that tasks are independent without precedence constraints or differential priorities. These assumptions effectively define the scope and boundaries of the specific challenge addressed by the methodologies presented subsequently.

\section{Methodology}
\subsection{Task Allocation via Enhanced Genetic Algorithm (EGA)}
The task allocation problem is modeled as a combinatorial optimization problem, where a set of medical rescue tasks $\mathcal{T} = \{T_1, T_2, \dots, T_M\}$ must be assigned to a fleet of heterogeneous vehicles $\mathcal{V} = \{V_1, V_2, \dots, V_N\}$. Each vehicle $V_i$ is constrained by its maximum operational range and capacity, and each task must be serviced exactly once. The objective is to minimize the total mission cost while satisfying all operational constraints.

GAs are a class of evolutionary algorithms inspired by natural selection and genetics. They are particularly well-suited for complex combinatorial optimization problems like ours, where the search space is vast and discontinuous. A GA works by maintaining a population of candidate solutions (individuals), which evolve over successive generations through processes of selection, crossover, and mutation. High-fitness solutions (those with lower mission cost) are more likely to be selected for reproduction, leading the population to converge towards optimal or near-optimal solutions.

To solve this optimization problem, we propose an EGA that builds upon the standard GA framework with several key improvements to enhance convergence speed and solution quality. The optimization objective is formulated as:
\begin{equation}
\min_{X} \sum_{i=1}^{N} \left( \sum_{j=0}^{n_i-1} d(P_{ij}, P_{i(j+1)}) \right),
\end{equation}
where $X$ denotes a complete task assignment solution, $P_{ij}$ represents the location of the $j$-th task assigned to vehicle $V_i$ (with $P_{i0}$ denoting the base location), and $d(\cdot, \cdot)$ is the Euclidean distance between two points.

Each solution $X$ must satisfy:
\begin{equation}
    \bigcup_{i=1}^{N} T_i = T, \quad T_i \cap T_j = \emptyset \quad \forall i \neq j, \quad i,j \in \{1, \dots, N\}, 
\end{equation}
where $\mathcal{T}_i$ is the subset of tasks assigned to vehicle $V_i$. This ensures that each task is uniquely and exclusively assigned.

To illustrate this encoding, consider a scenario with 3 vehicles (2 UAVs: $U_1$, $U_2$; 1 UGV: $G_1$) and 7 tasks ($T_1$ to $T_7$). A possible individual (chromosome) could be represented as: $\text{Chromosome: } [U_1: T_1, T_3, T_5 \mid U_2: T_2, T_6 \mid G_1: T_4, T_7]$.

In this example, the chromosome is a single sequence divided by vehicle identifiers. $U_1$ is assigned the task sequence $T_1 \rightarrow T_3 \rightarrow T_5$, $U_2$ is assigned $T_2 \rightarrow T_6$, and $G_1$ is assigned $T_4 \rightarrow T_7$. This encoding explicitly captures both the assignment of tasks to vehicles and the sequence in which those tasks are visited.

The EGA is designed as follows: Each individual in the GA population is represented as a sequence of task indices, grouped by the vehicle to which they are assigned. The fitness of an individual is evaluated according to the total path cost defined by the objective function (5). A tournament selection operator is employed to favor individuals with lower total cost. A partially matched crossover (PMX) operator is used to exchange assigned tasks between two parent individuals, followed by a repair mechanism to ensure feasibility. Mutation is performed by randomly swapping tasks between different vehicles or rearranging the local task order within a vehicle's sequence.

The enhancements in our EGA are two-fold:

1. Adaptive Mutation Probability: To balance exploration in early generations with exploitation in later generations, the mutation probability $\mu$ is dynamically adjusted over time:
    \begin{equation}
    \mu(g) = \mu_0 \times \exp\left(-\alpha \frac{g}{G}\right), 
    \end{equation}
    where $\mu_0$ is the initial mutation probability, $g$ is the current generation number, $G$ is the total number of generations, and $\alpha$ controls the decay rate. This allows for aggressive exploration initially and fine-tuning later.

2.  Elite Preservation: An elite preservation mechanism ensures that the top $E\%$ of individuals with the lowest cost are directly carried over to the next generation. This prevents the loss of high-quality solutions and accelerates convergence.

The overall EGA procedure iteratively improves the population until a stopping criterion is met. The pseudo-code for the proposed EGA is presented in Algorithm 1.

\begin{algorithm}[H]
\footnotesize
\caption*{\textbf{Algorithm 1} Enhanced Genetic Algorithm for Task Allocation}
\begin{algorithmic}
\REQUIRE Task set $\mathcal{T}$, Vehicle set $\mathcal{V}$, Population size $P$, Max generations $G$, Initial mutation rate $\mu_0$, Elite ratio $E$
\ENSURE Optimal task assignment $X_{\text{best}}$
\STATE Initialize population $P_{\text{pop}}$ with $P$ random individuals
\STATE Evaluate fitness for all individuals in $P_{\text{pop}}$
\STATE $X_{\text{best}} \gets$ best individual from $P_{\text{pop}}$
\FOR{$g = 1$ to $G$}
    \STATE Select parents from $P_{\text{pop}}$ using tournament selection
    \STATE Perform PMX crossover to generate offspring
    \STATE Apply repair mechanism to ensure task uniqueness
    \STATE Calculate mutation probability $\mu(g)$ using Eq. (7)
    \STATE Perform mutation on offspring with probability $\mu(g)$
    \STATE Evaluate fitness for new offspring
    \STATE Form new population: Combine new offspring with $E\%$ elites from $P_{\text{pop}}$
    \STATE $P_{\text{pop}} \gets$ new population
    \STATE Update $X_{\text{best}}$ if a better individual is found
\ENDFOR
\RETURN $X_{\text{best}}$
\end{algorithmic}
\end{algorithm}

\subsection{Trajectory Generation via Informed-RRT*}
After task allocation is completed, each unmanned vehicle must generate feasible trajectories to sequentially visit its assigned task points while avoiding obstacles. To efficiently plan paths in the cluttered post-disaster environment, we employ the Informed-RRT* algorithm, an improved sampling-based method that focuses the search on promising regions, accelerating convergence toward near-optimal paths.
Given a start position $P_{\text{start}}$ and a goal position $P_{\text{goal}}$, the path planning problem aims to find a collision-free trajectory $\mathcal{P}$ such that:
\begin{equation}
    \mathcal{P} = \{P_0, P_1, \dots, P_k\}, \quad P_0 = P_{\text{start}}, \quad P_k = P_{\text{goal}},
\end{equation}
while satisfying:
$P_i \notin \mathcal{O}, \quad \forall i = 0, \dots, k$, where $\mathcal{O}$ represents the set of obstacle regions.

Informed-RRT* improves upon classical RRT* by restricting sampling to an ellipsoidal subset of the space once an initial feasible path is found. The sampling domain is dynamically updated based on the current best path cost $c_{\text{best}}$ and the heuristic minimum cost $c_{\text{min}}$ between start and goal:
\begin{equation}
    \mathcal{X}_{\text{informed}} = \{ x \in \mathbb{R}^2 \mid \|x - x_{\text{start}}\| + \|x - x_{\text{goal}}\| \leq c_{\text{best}} \},
\end{equation}
thus significantly reducing unnecessary exploration and focusing computational effort where better solutions are likely to be found.

For each segment between consecutive tasks, a local path is planned using Informed-RRT*, and the full mission path for each vehicle is obtained by concatenating these local paths. The total trajectory cost for vehicle $V_i$ is computed as:
\begin{equation}
    C_i = \sum_{j=0}^{n_i-1} c(P_{ij}, P_{i(j+1)}),
\end{equation}
where $c(\cdot, \cdot)$ denotes the cost (typically Euclidean distance) between two consecutive planned waypoints.

The core of the Informed-RRT* algorithm lies in its dynamically defined sampling domain. Once an initial feasible path with cost $c_{\text{best}}$ is found, the sampling is restricted to an ellipsoidal region $X_{\text{informed}}$, mathematically defined as:
\begin{equation}
    X_{\text{informed}} = \{ \mathbf{x} \in \mathbb{R}^2 \mid \|\mathbf{x} - \mathbf{x}_{\text{start}}\| + \|\mathbf{x} - \mathbf{x}_{\text{goal}}\| \leq c_{\text{best}} \}
\end{equation}
where $\mathbf{x}_{\text{start}}$ and $\mathbf{x}_{\text{goal}}$ are the start and goal positions, respectively. This constraint ensures that only states potentially leading to a path with a cost lower than $c_{\text{best}}$ are sampled, which is the key mechanism for its accelerated convergence.

The use of Informed-RRT* provides significant advantages over the standard RRT*. By focusing the sampling effort within the informed set $X_{\text{informed}}$, the algorithm drastically reduces the number of samples wasted in regions of the state space that cannot yield an improved solution. This leads to a much faster convergence rate towards a near-optimal path. Furthermore, like its predecessor, Informed-RRT* retains the theoretical property of asymptotic optimality under certain conditions, guaranteeing that the solution cost will converge to the global optimum with probability one given sufficient time, providing a strong theoretical foundation for the trajectory generation phase.

To enhance the quality and practicality of generated trajectories, a post-processing stage is applied to remove redundant waypoints and slightly smooth the paths without significantly altering obstacle avoidance guarantees. This ensures that the final trajectories are not only feasible but also suitable for real-world execution by UAVs and UGVs.
Through the use of Informed-RRT*, the framework achieves efficient, collision-free trajectory generation even in complex, obstacle-rich disaster environments, forming a solid foundation for subsequent trajectory optimization.

\subsection{Trajectory Optimization with CMA-ES}
After generating feasible trajectories connecting the assigned task points, further optimization is necessary to improve overall mission efficiency. In particular, the visiting order of the assigned tasks significantly influences the total distance traveled and, consequently, the mission completion time and energy consumption. Therefore, a trajectory optimization phase is introduced to refine the task execution sequence for each vehicle. To solve this permutation optimization problem, the Covariance Matrix Adaptation Evolution Strategy (CMA-ES) is employed.

The trajectory optimization problem for a vehicle $V_i$ can be formulated as finding the optimal task visiting sequence $\pi_i$ that minimizes the total travel cost:
\begin{equation}
    \min_{\pi_i} C_i(\pi_i) = \sum_{j=0}^{n_i-1} d\big(P_{\pi_i(j)}, P_{\pi_i(j+1)}\big),
\end{equation}
where 
$n_i$ denotes the number of tasks assigned to vehicle $V_i$, 
$\pi_i$ is a permutation of the task indices that specifies the visiting order, 
with $\pi_i(j)$ indicating the index of the $j$-th task visited by $V_i$. 
The 2D coordinate of the $\pi_i(j)$-th task is denoted by $P_{\pi_i(j)}$, 
and we define $P_{\pi_i(0)} = P_{\pi_i(n_i)} = B$, where $B$ is the base location (depot), 
ensuring that each vehicle starts and ends its route at the depot. 
The function $d(\cdot, \cdot)$ computes the Euclidean distance between two points in the plane. 
Finally, $C_i(\pi_i)$ represents the \textbf{total path cost} for vehicle $V_i$, 
defined as the sum of the travel distances along the route $\pi_i$, 
i.e., the total distance traveled by $V_i$ when executing the task sequence $\pi_i$. 
This cost is the objective function minimized by the CMA-ES algorithm.

Since the optimization domain is discrete (permutations of tasks), but CMA-ES operates in a continuous space, a continuous relaxation approach is adopted. Each candidate solution is represented as a real-valued vector, where the ranking of elements determines the visiting order. During evaluation, the vector elements are sorted to yield a specific permutation $\pi_i$.

CMA-ES iteratively updates a multivariate normal distribution over the solution space by adapting its covariance matrix, allowing it to efficiently explore and exploit the search space. In each generation, a set of candidate solutions is sampled, evaluated using the cost function $C_i(\pi_i)$, and used to update the distribution parameters. The evolution of the mean vector $\mathbf{m}$ and covariance matrix $\mathbf{C}$ follows:
\begin{equation}
    \mathbf{m}_{g+1} = \sum_{k=1}^{\lambda} w_k \, \mathbf{x}_k^{(g)},
\end{equation}
\begin{equation}
    \mathbf{C}_{g+1} = (1 - c_1 - c_{\mu}) \, \mathbf{C}_g + c_1 \mathbf{p}_c \mathbf{p}_c^T +
\end{equation}
     \[c_{\mu} \sum_{k=1}^{\mu} w_k \, (\mathbf{x}_k^{(g)} - \mathbf{m}_g)(\mathbf{x}_k^{(g)} - \mathbf{m}_g)^T,\]
where $\mathbf{x}_k^{(g)}$ are selected offspring, $w_k$ are recombination weights, and $c_1, c_{\mu}$ are learning rates for rank-one and rank-$\mu$ updates, respectively.

CMA-ES is particularly well-suited for the trajectory sequence optimization problem due to its powerful adaptive mechanism. Unlike algorithms with fixed search strategies, CMA-ES learns the structure of the optimization landscape by continuously adapting its covariance matrix $C$. This allows it to efficiently identify and exploit correlations between decision variables (i.e., task positions in the sequence), effectively performing a principled local search in promising directions. Its ability to handle non-linear, non-convex, and potentially noisy objective functions makes it robust for refining paths in complex environments. The continuous relaxation approach bridges the gap between the discrete nature of the TSP-like problem and the continuous optimization capability of CMA-ES, enabling a sophisticated search that can escape local minima more effectively than simpler heuristics, ultimately leading to smoother and more efficient final trajectories.

The CMA-ES process continues until convergence criteria are met, such as a maximum number of generations or stagnation of fitness improvement. The final optimized visiting order $\pi_i^*$ is then used to reconstruct the full vehicle trajectory.
By applying CMA-ES to optimize task visiting sequences, the proposed framework effectively reduces redundant movements, shortens total mission time, and produces smoother, more efficient trajectories, enhancing the overall performance of the UAV-UGV cooperative rescue system.

\subsection{Integrated Cooperative Planning Framework}


The integrated planning framework commences by modeling the post-disaster environment, defining obstacle regions $O$, task points $T$, and the base station $B$. Subsequently, an EGA is employed to allocate medical rescue tasks $T$ among the heterogeneous fleet of UAVs $U$ and UGVs $G$, aiming to optimize overall mission efficiency. Following task allocation, the Informed-RRT* algorithm generates initial collision-free and constraint-compliant trajectories $P_i(t)$ for each vehicle $i$ to visit its assigned tasks. CMA-ES is then utilized to optimize the task visiting sequence $\pi_i$ and refine the trajectory for each vehicle, minimizing the total mission completion time $T_{\text{total}}$ and the sum of path costs $\sum C_i(\pi_i)$ (which directly relates to energy consumption $\sum E_i$). Finally, the complete plan is evaluated to ensure adherence to all constraints, including safety distances $d_{\text{safe}}$, dynamics, energy limits $E_i$, and full task coverage.Fig. 2 shows the integration framework diagram, showing the connections between the modules.

\begin{figure}[H]
    \centering
    \includegraphics[width=0.9\linewidth]{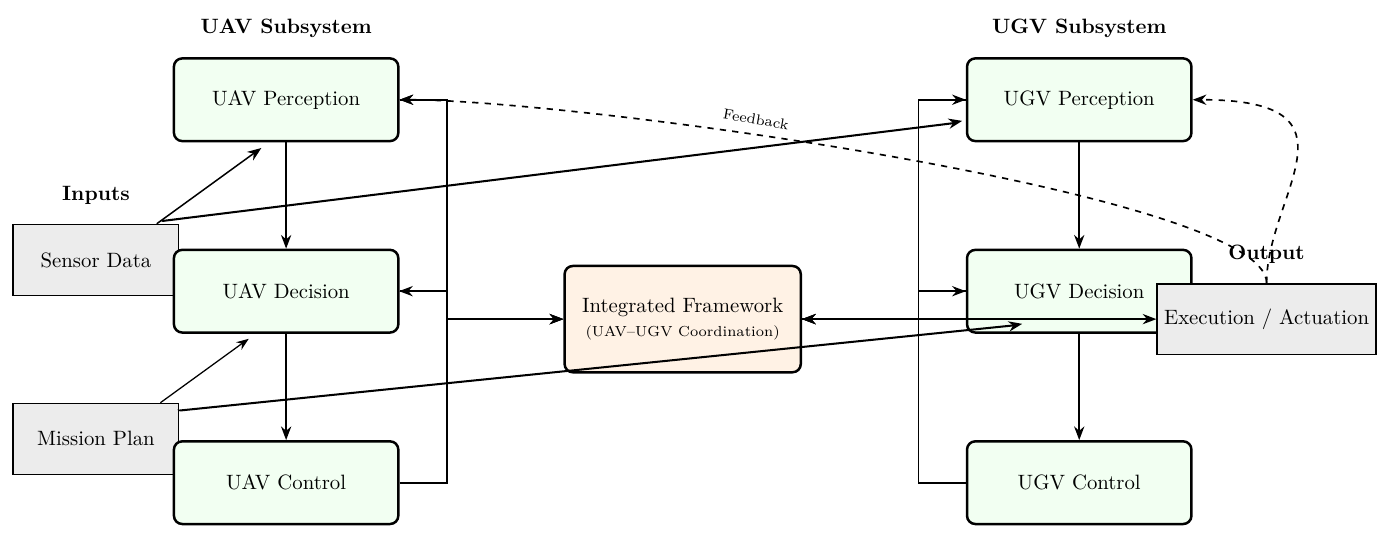}
    \caption{Integrated framework diagram }
    \label{fig:enter-label}
\end{figure}

The strength of this framework lies in its structured integration of specialized algorithms. The EGA provides a robust global task distribution considering vehicle heterogeneity. Its output (task assignments $X$) directly defines the subproblems solved by the subsequent path planning and sequence optimization stages for each vehicle. Informed-RRT* acts as a crucial feasibility engine, ensuring that generated paths respect environmental constraints ($O$) and basic vehicle limitations. The path costs derived from RRT* can potentially inform the initial cost estimates used in EGA, although this feedback loop might be simplified in practice. CMA-ES then performs a local optimization per vehicle, refining the solution by optimizing the task execution order based on travel costs derived from the environment geometry (and potentially informed by RRT*'s findings about connectivity). This sequential process, where global allocation precedes local path generation and sequence refinement, allows the framework to handle the high complexity of the combined problem in a tractable manner.


The framework explicitly supports UAV-UGV cooperation by design. The EGA module inherently considers the distinct capabilities (speed, range, payload) encoded in the vehicle parameters during the task allocation phase, assigning tasks to the most suitable vehicle type. This allows the system to exploit the complementary strengths noted earlier–UAVs for rapid response and access, and UGVs for endurance and heavy loads. Regarding inter-vehicle coordination in the shared workspace, the current framework primarily relies on the Informed-RRT* algorithm's ability to generate paths avoiding static obstacles. Dynamic inter-vehicle collision avoidance (deconfliction) is managed implicitly by ensuring sufficient spatial or temporal separation, potentially requiring downstream scheduling or velocity adjustments by lower-level controllers. Explicit dynamic deconfliction within this high-level planning framework is considered beyond the current scope but represents an important area for future extension.

This framework capitalizes on the complementary strengths of UAVs and UGVs. UAVs, with their high mobility and speed, can rapidly traverse areas and overcome moderate obstacles, making them suitable for urgent deliveries and aerial reconnaissance. Conversely, UGVs offer higher payload capacity and longer endurance, ideal for transporting heavier supplies or tasks requiring ground presence. The EGA-based task allocation mechanism intelligently assigns tasks by considering these distinct capabilities, ensuring that tasks are allocated to the most appropriate vehicle type to maximize overall system efficiency. Cooperative planning ensures safe and efficient operation in a shared environment, enabling the heterogeneous team to tackle complex rescue missions effectively.

This integrated approach offers several advantages for the target application. By separating concerns—global allocation, feasible pathfinding, and local sequence optimization decomposes the complex overall problem into manageable subproblems. It systematically leverages the strengths of different optimization paradigms suited for different aspects of the problem . This structured integration facilitates the generation of globally coordinated, locally optimized, feasible, and efficient plans for heterogeneous multi-robot teams operating in challenging post-disaster environments, ultimately aiming to improve the timeliness and effectiveness of medical rescue operations.

\section{Simulation Results}
\subsection{Simulation Setup}

To validate the framework, simulations are conducted in a realistic post-disaster environment. The heterogeneous fleet includes UAVs and UGVs with distinct operational limits. All vehicles must adhere to their dynamic constraints, energy limits ($E_i$), and maintain a minimum safety distance $d_{\text{safe}}$ from obstacles and other vehicles. The energy constraint for each vehicle is enforced as:
\begin{equation}
C_i(\pi_i) \leq E_i,
\end{equation}
where $C_i(\pi_i)$ is the total path cost (in distance units) and $E_i$ is the maximum allowable cost (converted from energy budget via a vehicle-specific efficiency model). This ensures that longer paths, which consume more energy, are penalized during planning.

The disaster environment is modeled as a 20 km × 20 km two-dimensional area representing a severely impacted urban zone following a major earthquake. The terrain is assumed to be cluttered with collapsed buildings, rubble fields, and inaccessible zones, creating a complex navigation landscape for rescue operations. Static obstacles are embedded within the environment to simulate real-world hazards that UAVs and UGVs must avoid during mission execution.

Within the disaster area, five major obstacle regions are defined, each modeled as a circular impassable area characterized by a center coordinate and a radius. The obstacles are randomly distributed across the map but strategically positioned to challenge both aerial and ground navigation. The size of obstacles varies between 500 meters and 1.5 kilometers in diameter, representing collapsed building complexes or large debris zones.

A total of 15 task points are randomly scattered across the environment to simulate emergency medical rescue missions, such as supply deliveries or on-site first aid. Each task point $T_j$ is uniquely identified by a coordinate:
$T_j = (x_j, y_j), \quad j = 1, 2, \dots, 15$.
All unmanned vehicles (both UAVs and UGVs) begin and end their missions at a centralized base station located at the center of the environment, specifically at coordinates:
$B = (0, 0)$. 
The centralized base serves as the operational hub for task allocation, mission monitoring, and vehicle recovery.

For the simulation experiments, a heterogeneous fleet comprising ten UAVs, denoted as the set $\mathcal{U}$, and five UGVs, forming the set $\mathcal{G}$, was configured to represent the available rescue assets. The operational characteristics defining these distinct vehicle types were specified to reflect realistic capabilities and limitations relevant to post-disaster scenarios. The UAVs were modeled with a maximum speed of 60 km/h and a maximum operational range of 15 km, reflecting typical endurance based on battery capacity (potentially equivalent to approximately 20 km of ideal flight). Each UAV possessed a payload capacity limited to 5 kg, suitable for delivering smaller, critical medical packages rapidly. In contrast, the UGVs were characterized by a lower maximum speed of 30 km/h but offered significantly greater endurance with a maximum operational range of 25 km (potentially equivalent to 30 km of driving range under ideal conditions) and a substantially larger payload capacity of 50 kg. This higher payload makes them well-suited for transporting bulkier supplies or potentially assisting in casualty evacuation tasks. Both UAVs and UGVs were assumed to be equipped with autonomous obstacle avoidance capabilities, enabling navigation within the complex, obstacle-ridden simulated environment. This heterogeneous composition allows the system to leverage the speed and aerial access of UAVs alongside the endurance and carrying capacity of UGVs.

All vehicles must adhere to their respective dynamic limits, energy constraints ($C(P_i) \le E_i$), and maintain minimum safety distances ($d_{\text{safe}}$) from obstacles and other vehicles during operation.

In the simulation implementation, we configured the parameters of the optimization algorithms used. For the EGA for task allocation, its key parameters (such as population size, number of iterations, crossover and mutation probabilities, etc.) are set according to the specific scale and complexity of the problem being studied. In order to improve the algorithm's search efficiency and the quality of the solution, we may also incorporate mechanisms such as adaptive mutation rate adjustment and elite retention strategies. Similarly, for the CMA-ES for trajectory sequence optimization, the parameters required for its operation, such as the initial mean vector, covariance matrix, and population (sample) size, are also carefully selected and configured based on the characteristics of the optimization landscape of the optimization objective function.

The primary metrics used to evaluate the performance of the proposed framework include:
1.Total Mission Completion Time ($T_{\text{total}}$): Time until the last vehicle returns to base.
2.Total Path Length ($\sum L_i$): Sum of all vehicle trajectories. For clarity, $L_i \equiv C_i(\pi_i)$, so $\sum L_i = \sum C_i(\pi_i)$.
3.Energy Consumption Estimation ($\sum E_i$): Estimated as $\sum E_i = \sum \eta_i \cdot C_i(\pi_i)$, where $\eta_i$ is the energy consumption rate (e.g., J/m) for vehicle $V_i$. Thus, energy is directly proportional to the optimized path cost $C_i(\pi_i)$.











\subsection{Results and Analysis}

\begin{figure}[H]
    \centering
    \includegraphics[width=0.9\linewidth]{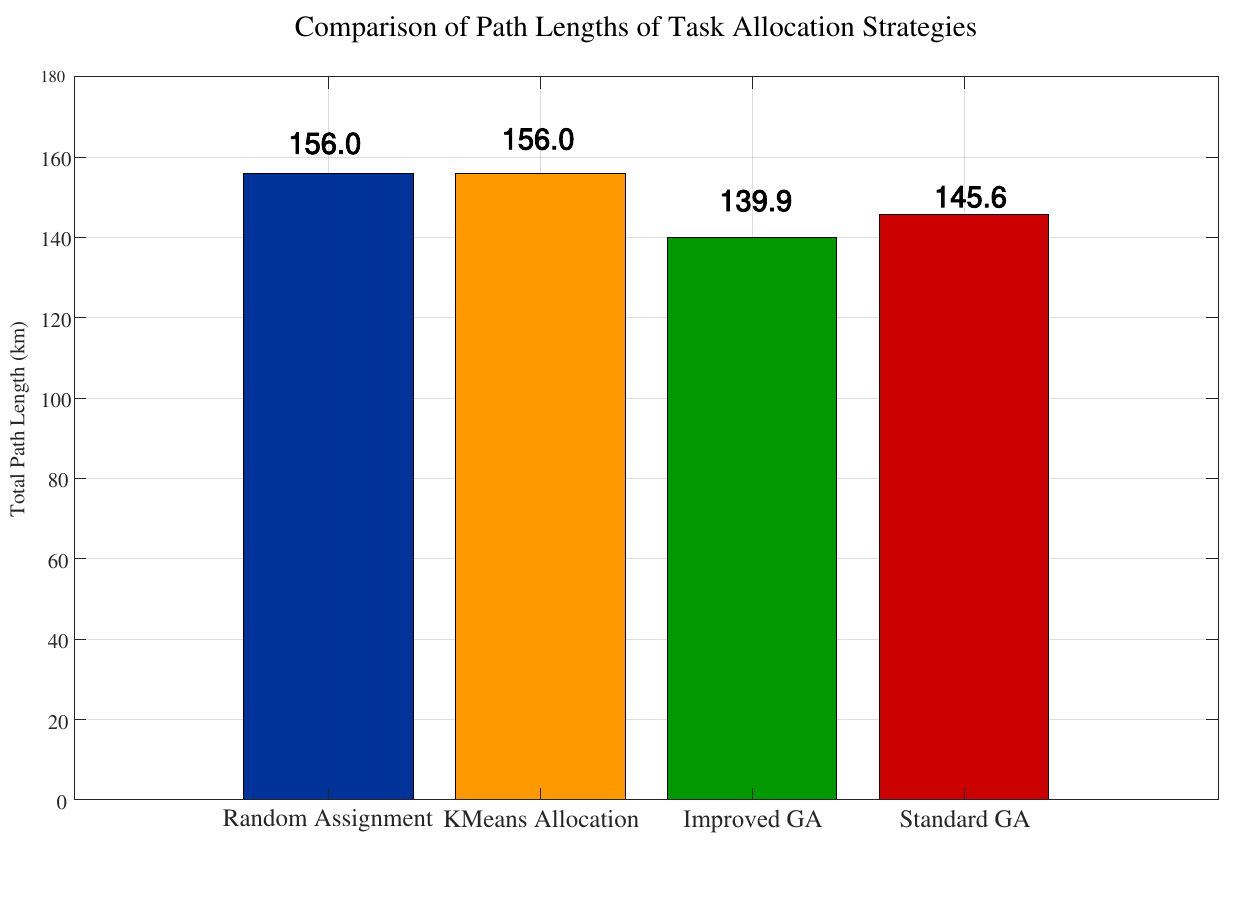}
    \caption{ Comparison of total path length generated by different task allocation strategies. The proposed Enhanced Genetic Algorithm (EGA) achieves the shortest total path length, demonstrating its superior efficiency in optimizing the overall mission route compared to Random Assignment, K-Means clustering and standard GA.}
    \label{fig:enter-label}
\end{figure}
Comparative experiments were designed to evaluate the EGA-based task allocation against baseline methods such as random assignment, K-Means clustering, improved GA and standard GA optimization allocation. Expected results, visualized using bar charts Fig. 3, will demonstrate that the EGA significantly reduces the total path length and/or mission completion time compared to these simpler strategies, highlighting its effectiveness in finding efficient assignments under complex constraints.

Fig. 4 shows the convergence behavior of the GA during the optimization process. A sharp reduction in the Best Total Distance occurs very early, within the first 1-2 generations. Subsequently, the convergence is characterized by periods of plateau interspersed with significant decreases, particularly noticeable around generation 42, and a more substantial drop occurring between generations 81 and 85. The curve eventually stabilizes after approximately generation 92, indicating that the algorithm has converged to a near-optimal solution. This progression demonstrates the algorithm's search process, finding substantial improvements in distinct phases.

A key aspect of the proposed framework is the trajectory optimization performed using CMA-ES, which refines the sequence of visiting assigned tasks for each vehicle. To evaluate the effectiveness of this optimization step, a comparative analysis was conducted. We compared the efficiency of the final paths, generated after applying CMA-ES sequence optimization, against the initial feasible paths generated directly by Informed-RRT* based on an unoptimized task sequence.

\begin{figure}[H]
    \centering
    \includegraphics[width=0.7\linewidth]{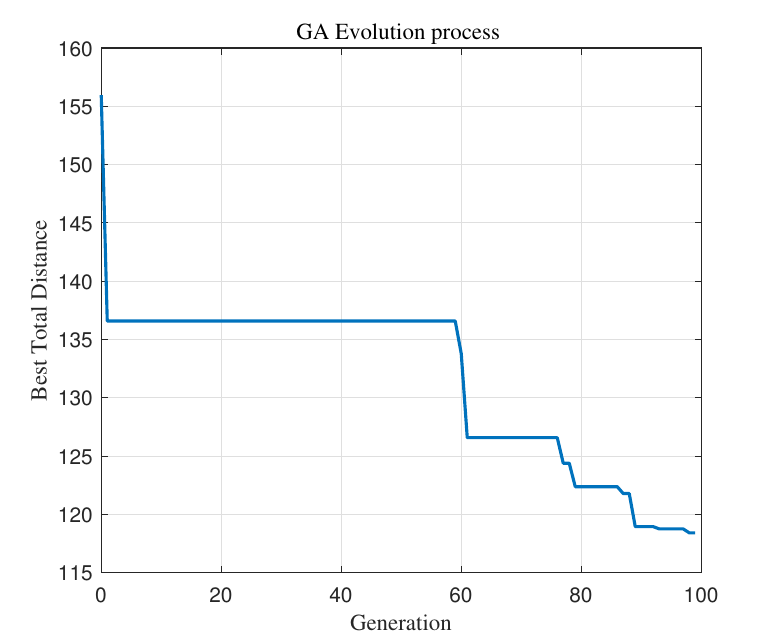}
    \caption{Convergence curve of the EGA for task allocation. The algorithm rapidly reduces the best total path length in the early generations and converges to a near-optimal solution after approximately 92 generations, indicating stable performance. }
    \label{fig:enter-label}
\end{figure}

Quantitative results of this comparison are presented in Fig. 5. This figure directly contrasts a key performance metric, such as total path length, before and after CMA-ES optimization. As indicated by the data shown, the application of CMA-ES leads to a substantial reduction in this metric, demonstrating a significant improvement in path length efficiency solely due to optimizing the task visiting order.

Complementing this quantitative summary, Fig. 6 provides a qualitative and potentially more detailed quantitative perspective. Fig. 6 presents visual trajectory plots alongside further numerical comparisons. These plots would visually depict representative paths for vehicles before sequence optimization and the corresponding paths after CMA-ES optimization, which are expected to appear more direct and efficient in connecting the assigned task locations while still avoiding obstacles. The accompanying quantitative data in Fig. 6 would further reinforce the findings from Fig. 5, possibly breaking down the distance savings per vehicle or providing other relevant efficiency metrics.

Fig. 5 and 6 provide compelling evidence for the value of the CMA-ES trajectory optimization stage. By intelligently reordering the sequence in which tasks are visited, CMA-ES effectively shortens the required travel distances and streamlines the routes, thereby significantly improving the overall efficiency and likely reducing the energy consumption of the planned missions compared to using only the initial feasible paths from Informed-RRT*. This improvement stems from the powerful adaptive search capability of CMA-ES. By iteratively updating the covariance matrix $C$, CMA-ES can learn the correlations among task points. For instance, in a dense task region, the algorithm learns that tasks within this area tend to be visited consecutively, thereby focusing the search toward solution spaces that generate "clustered" visiting sequences. This adaptive mechanism, based on historical search information, enables the algorithm to efficiently explore and converge to globally optimal or near-optimal visiting orders, rather than merely performing local refinements, thus achieving a significant reduction in path length.

\begin{figure}[H]
    \centering
    \includegraphics[width=0.8\linewidth]{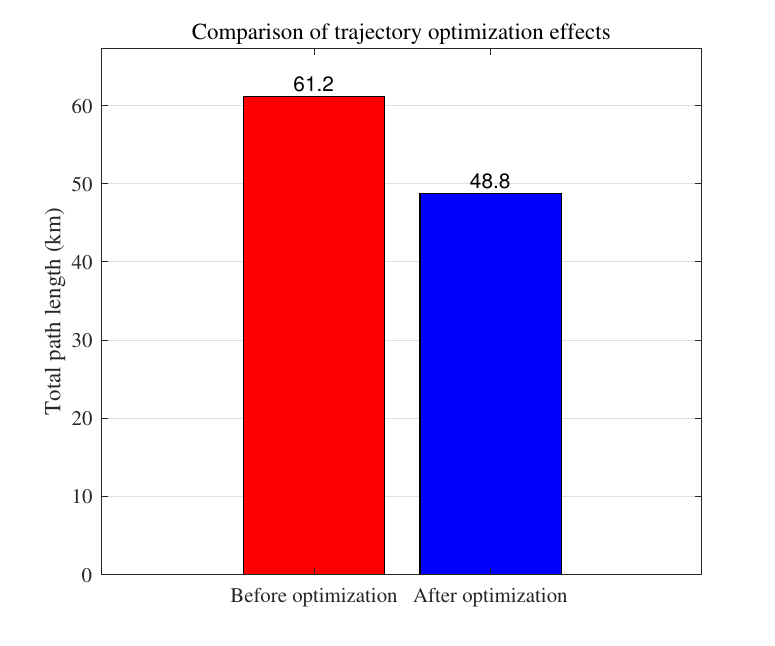}
    \caption{ Quantitative comparison of total path length before and after CMA-ES trajectory optimization. The application of CMA-ES reduces the total path length from 61.2 km to 48.8 km, achieving a 20.3\% improvement in route efficiency for this specific mission scenario.}
    \label{fig:enter-label}
\end{figure}

The analysis will confirm that all trajectories generated by the integrated framework are collision-free, respecting obstacle boundaries and inter-vehicle safety distance while adhering to vehicle operational constraints like maximum range. The optimization process contributes not only to shorter paths but potentially also to smoother trajectories suitable for practical execution.


The simulation results effectively underscore the benefits derived from employing a heterogeneous UAV-UGV team coordinated by the proposed framework. The significantly lower mission completion times achieved by the GA-based optimization compared to K-Means and Random Allocation methods, as detailed in Table 1, strongly suggest that the intelligent task allocation mechanism successfully leverages the distinct advantages of each vehicle type. By considering factors implicitly related to capabilities—such as assigning speed-critical or obstacle-sensitive tasks potentially better suited to UAVs, while allocating high-payload or long-range tasks to UGVs—the framework achieves superior overall mission performance compared to approaches less sensitive to vehicle heterogeneity. In contrast, K-Means and Random, both 100 min in the baseline.
\begin{table}[H]
\centering
\caption{Extensibility Analysis}
\label{tab:5 drones}
\hspace*{-0.2cm}
\scalebox{0.55}{
\renewcommand{\arraystretch}{2} 
\begin{tabular}{ccccc}
\hline
\textbf{Number of Tasks} & \textbf{Number of Vehicles} & \textbf{GA optimization} & \textbf{K-Means} & \textbf{Random Allocation} \\ \hline
15       & \textbf{10U+5G}            & \textbf{26.7 min}             & \textbf{100.0 min} & \textbf{100.0 min}               \\
30       & \textbf{10U+5G}            & \textbf{39.8 min}             & \textbf{104.5 min} & \textbf{68.2 min}                \\
60       & \textbf{20U+10G}           & \textbf{35.0 min}             & \textbf{116.3 min} & \textbf{133.2 min}               \\ \hline
\end{tabular}
}
\end{table}

\begin{figure}[H]
    \centering
    \includegraphics[width=0.9\linewidth]{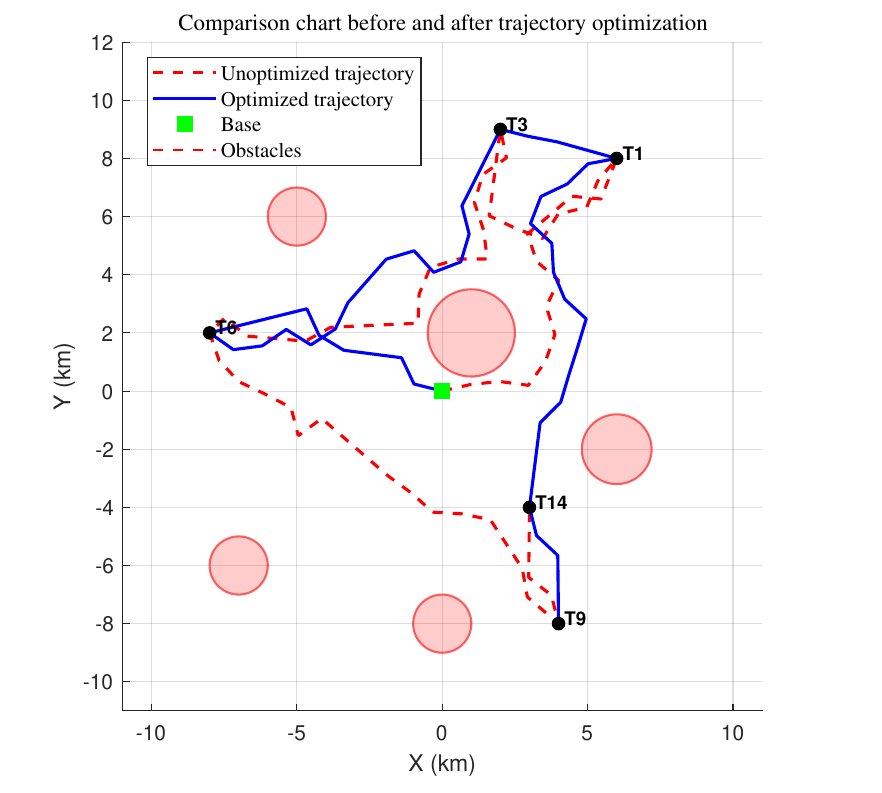}
    \caption{Visual comparison of a representative vehicle trajectory before and after CMA-ES optimization. The optimized trajectory (blue) is significantly shorter and more direct than the unoptimized one (red), while both paths successfully avoid obstacles (gray shaded areas) and connect the base station and assigned task points.}
    \label{fig:enter-label}
\end{figure}
Furthermore, the framework's performance scalability was investigated by increasing the number of tasks and vehicles. The experimental results presented in Table 1 validate the hypothesis that the proposed methods demonstrate good scalability. When the number of tasks was doubled from 15 to 30 while keeping vehicle numbers constant, the mission completion time for the GA-optimized approach increased moderately (from 26.7 min to 39.8 min), maintaining a significant advantage over other methods. More notably, when both tasks and vehicles were substantially increased (60 tasks, 20U+10G), the GA optimization achieved a completion time (35.0 min) that was not only faster than the 30-task scenario but also comparable to the original baseline, indicating highly effective parallelization and resource utilization. This contrasts sharply with the performance degradation observed for K-Means and Random Allocation under increased scale, confirming the proposed integrated framework maintains efficiency well even when dealing with larger, more complex scenarios.
\begin{figure}[H]
    \centering
    \includegraphics[width=1\linewidth]{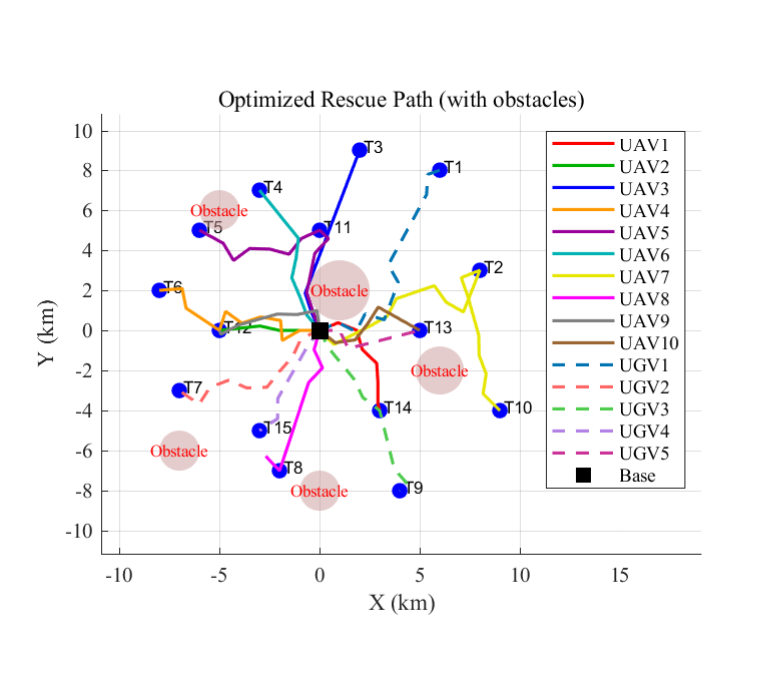}
    \caption{Final cooperative mission plan for the heterogeneous UAV-UGV fleet in the simulated disaster environment. The trajectories of UAVs (solid lines) and UGVs (dashed lines) originate from the central base, visit their assigned task locations, and return to the base, while navigating around obstacle regions. This demonstrates the framework's ability to generate a complete, feasible, and coordinated plan.}
    \label{fig:enter-label}
\end{figure}

To provide a visual representation of the complete mission plan generated by the integrated framework, Fig. 7 displays the final path trajectories for the heterogeneous fleet of UAVs and UGVs operating within the simulated post-disaster environment. This figure overlays the calculated paths onto the environment map, clearly showing the locations of the base station, task points, and obstacle regions. Distinct visual styles (e.g., different colors or line types) are used to differentiate the trajectories of UAVs from those of UGVs. The depicted paths illustrate the practical output of the entire planning process: vehicles originating from the base, sequentially visiting their assigned task locations as determined by the EGA and optimized by CMA-ES, and subsequently returning to the base. Critically, Fig. 7 serves to visually verify the feasibility and effectiveness of the Informed-RRT* path generation and the overall framework, demonstrating that the generated trajectories successfully navigate around obstacles, thereby ensuring collision-free operation. Furthermore, it visually highlights the cooperative nature of the mission, potentially showing UAVs taking more direct routes over certain terrain while UGVs follow ground-based paths, reflecting the successful integration and utilization of the heterogeneous fleet's complementary capabilities.

In conclusion, the extensive simulation experiments detailed in this chapter provide substantial validation for the proposed integrated cooperative planning framework. The results consistently demonstrate the effectiveness of combining the Enhanced Genetic Algorithm for task allocation, Informed-RRT* for path generation, and CMA-ES for sequence optimization. Quantitative comparisons confirmed significant improvements in mission efficiency, notably reduced completion times and total path lengths, compared to baseline allocation strategies. Furthermore, the analysis verified the framework's ability to generate feasible, collision-free trajectories and effectively leverage the complementary strengths of the heterogeneous UAV-UGV team. Critically, the scalability analysis indicated that the proposed approach maintains its efficiency even as the complexity of the scenario increases in terms of task and vehicle numbers. Collectively, these simulation findings underscore the potential and robustness of the framework for enhancing coordinated robotic response in challenging post-disaster medical rescue operations.











\subsubsection{Discussion on Results and System Limitations}
The simulation results presented in Figs. 3--7 and Table 1 consistently demonstrate the superior performance of the proposed integrated framework. The EGA effectively leverages vehicle heterogeneity, assigning tasks to UAVs for rapid response and to UGVs for heavy payloads, resulting in significantly reduced mission times compared to baseline methods. The CMA-ES optimization further refines these plans, achieving notable reductions in total path length and, by extension, energy consumption.

However, a critical analysis of the results also reveals several inherent limitations of the current framework, which are important to acknowledge:

\begin{enumerate}
    \item \textbf{Static Environment Assumption:} The framework assumes a static environment with fixed obstacles and tasks, neglecting real-world dynamics such as shifting debris, emerging tasks, or changing weather, and lacks real-time replanning capability.

    \item \textbf{Simplified Inter-Vehicle Coordination:} Dynamic collision avoidance is not explicitly handled in the high-level planner, relying instead on lower-level controllers, which may lead to conflicts in dense operations.

    \item \textbf{Idealized System Models:} Simulations assume perfect communication and ideal vehicle dynamics, without accounting for communication delays, packet loss, or control noise prevalent in disaster zones.

    \item \textbf{Abstracted Energy Model:} Energy consumption is approximated by path length ($C(P_i)$), omitting effects of speed, terrain, and environmental conditions on actual energy usage.
\end{enumerate}

These limitations, while necessary for the tractability of the initial problem, highlight the gap between the simulated environment and the complexities of real-world deployment. They also naturally point towards the most promising directions for future research, which are discussed in the following section.
\section{Conclusion}

This paper has proposed a cooperative trajectory optimization and task allocation framework for heterogeneous UAV-UGV teams in post-disaster medical rescue scenarios. By integrating an EGA for global task distribution, Informed-RRT* for local collision-free path generation, and CMA-ES for sequence refinement, the framework effectively addresses the challenges of complex, obstacle-ridden environments. Simulation results demonstrate that the proposed method significantly reduces total mission completion time and overall path length compared to baseline strategies, validating its effectiveness in improving rescue efficiency. The framework successfully leverages the complementary strengths of UAVs and UGVs, enabling intelligent task assignment and generating feasible, optimized trajectories. The systematic approach—separating global allocation, path feasibility, and local sequence optimization—provides a scalable and robust solution for coordinating heterogeneous robot teams.

Despite its promising results, this work has several limitations. The assumption of a static environment limits its applicability; future work will focus on developing a dynamic replanning mechanism for real-time response to new tasks and moving obstacles.

\section{Code Availability}
The source code for the UAV-UGV cooperative trajectory optimization and task allocation for medical rescue tasks in Post-Disaster environments presented in this paper is openly available on GitHub at:\url{https://github.com/Cherry0302/disaster_uav_ugv_rescue_planner}


\end{multicols}
\end{document}